# Linguistic trajectories of bipolar disorder on social media

Laurin Plank[1*] and Armin Zlomuzica[1]

[1]Department of Behavioral and Clinical Neuroscience, Ruhr-University Bochum (RUB), D-44787 Bochum, Germany.

[*]Correspondence should be addressed to: Laurin Plank. Department of Behavioral and Clinical Neuroscience, Ruhr-University Bochum (RUB), Massenbergstr- 9-13, D-44787 Bochum, Germany, e-mail: laurin.plank@ruhr-uni-bochum.de.

### Abstract

**Background:** Language use offers valuable insight into affective disorders such as bipolar disorder (BD), yet past research has been cross-sectional and limited in scale.

**Objective:** Here, we demonstrate that social media records can be leveraged to study longitudinal language change associated with BD on a large scale.

**Methods:** Using a novel method to infer diagnosis timelines from user self-reports, we compared users self-identifying with BD, depression, or no mental health condition.

**Results:** The onset of BD diagnosis corresponded with widespread linguistic shifts reflecting mood disturbance, psychiatric comorbidity, substance abuse, hospitalization, medical comorbidities, interpersonal concerns, unusual thought content, and altered linguistic coherence. In the years following the diagnosis, discussions of mood symptoms were found to fluctuate periodically with a dominant 12-month cycle consistent with seasonal mood variation.

**Conclusions:** These findings suggest that social media language captures linguistic and behavioral changes associated with BD and might serve as a valuable complement to traditional psychiatric cohort research.

## Introduction

Bipolar disorder (BD) is a severe and lifelong mental disorder characterized by episodes of (hypo)mania and depression (Clemente et al., 2015). The onset of BD presents a major incision into the lives of affected individuals which is reflected by high rates of functional impairment and a 9-fold increased suicide risk (Burdick et al., 2022; Crump et al., 2013; McIntyre et al., 2020). During mania, affected individuals experience expansive mood, increased energy, a decreased need for sleep, and engage in impulsive behavior (Association & others, 2013; McIntyre et al., 2020; Ramírez-Martín et al., 2020). Depressive episodes are instead marked by low mood, reduced energy, and an increased sleep duration (Association & others, 2013).

Next to traditional methods such as self-report questionnaires or behavioral observations, analyses of freely-expressed language have been used to gain insight into the behavioral aberrations that characterize BD (Harvey et al., 2022; Zaher et al., 2025). Language aberrations are a central part of the clinical presentation and diagnosis of BD (American Psychiatric Association, 2022; Zaher et al., 2025). During mania, patients may exhibit pressured speech that is characterized by “jokes, puns, and amusing theatricalities” (American Psychiatric Association, 2022; Zaher et al., 2025). Depressive episodes are marked by decreased speech amount or a lower variety of content (American Psychiatric Association, 2022; Zaher et al., 2025). Similar to what can be observed in psychosis (Bilgrami, 2022; Stein et al., 2025), language in BD may exhibit signs of formal thought disorder which is reflected by phenomena such as incoherence or derailment (Andreasen, 1986; Arslan et al., 2024; Stein et al., 2025; Zaher et al., 2025). Other linguistic changes are a heightened concern with certain topics such as power and achievement (Andreasen & Pfohl, 1976) as well as an increased use of specific syntactic elements such as personal pronouns (Andreasen & Pfohl, 1976; Arslan et al., 2024; Cohan et al., 2018; Lorenz & Cobb, 1952).

While manual ratings of language have predominated earlier studies, automated methods from the field of natural language processing (NLP) are increasingly being used to detect changes in the content and form of language in BD (Arslan et al., 2024; Mota et al., 2014; Palaniyappan et al., 2019). In psychosis, where many of these methods have been pioneered (Elvevåg et al., 2007), NLP-based classifiers can detect the presence or predict the onset of psychotic episodes with accuracies reaching as high as 80-90% (Bedi et al., 2015; Ciampelli et al., 2023; Corcoran et al., 2018). This provides a strong clinical motive for studying language aberrations in BD where they could also be leveraged for early symptom detection or relapse monitoring (Anmella et al., 2024, 2024; Zaher et al., 2025). However, as of yet only few studies have employed NLP methods to study language disturbance in the context of BD (Arslan et al., 2024; Mota et al., 2014; Palaniyappan et al., 2019).

Of the studies that have been conducted, most collected cross-sectional samples which do not permit insight into longitudinal linguistic change associated with BD (Zaher et al., 2025). A reason for this might be the prohibitive nature of performing large-scale longitudinal research (Dzogang et al., 2016). Due to its massive scale and longitudinal scope, data gathered from social media (SM) platforms is emerging as a valuable resource for research into mental health disorders (Plank & Zlomuzica, 2024b, 2025; Wongkoblap et al., 2017). Unlike data collected in laboratory settings, SM provides a lens into unobstructed real-life behavior (Plank & Zlomuzica, 2025; Wongkoblap et al., 2017).

Past studies using SM data of users with BD have investigated the expression of emotion and negative feelings (Yoo et al., 2019), discussions around health-related topics (Coppersmith et al., 2015; Low et al., 2020), hypersexuality (Harvey, Rayson, Lobban, Palmier-Claus, Dolman, Chataigné, et al., 2025; Harvey, Rayson, Lobban, Palmier-Claus, Dolman, & Jones, 2025), and built classifiers to differentiate between healthy SM users and those with BD (Coppersmith et al., 2015; Kim et al., 2020) (for a review, see (Harvey et al., 2022)). Although SM provides data across long periods of time, to the best of our knowledge, no study has yet studied SM records to investigate how language develops longitudinally in SM users with BD. We assume that the main reason for this has been methodological; information on the time of a user's BD diagnosis is not readily available in SM datasets.

This paper addressed this gap by proposing a novel method to extract diagnosis times from SM records. This allowed us to study linguistic changes that occur as users with BD move towards their diagnosis and revealed language change in the years following the diagnosis.

Fig. 1 gives an overview of the presented study. We used data from the popular SM platform Reddit and identified users who self-disclosed suffering from BD. We focused on aspects of language content (i.e., *what* is being said) and language form (i.e., *how* it is being said) as both of these domains have previously been shown to be altered in BD (Harvey et al., 2022; Zaher et al., 2025). NLP methods were used to study the verbosity, emotionality, semantic coherence, syntax, and topics of posts (see Post analysis). These changes were compared to those found in users with self-disclosed depression (UD) and those with no self-disclosed mood disorder (HC).

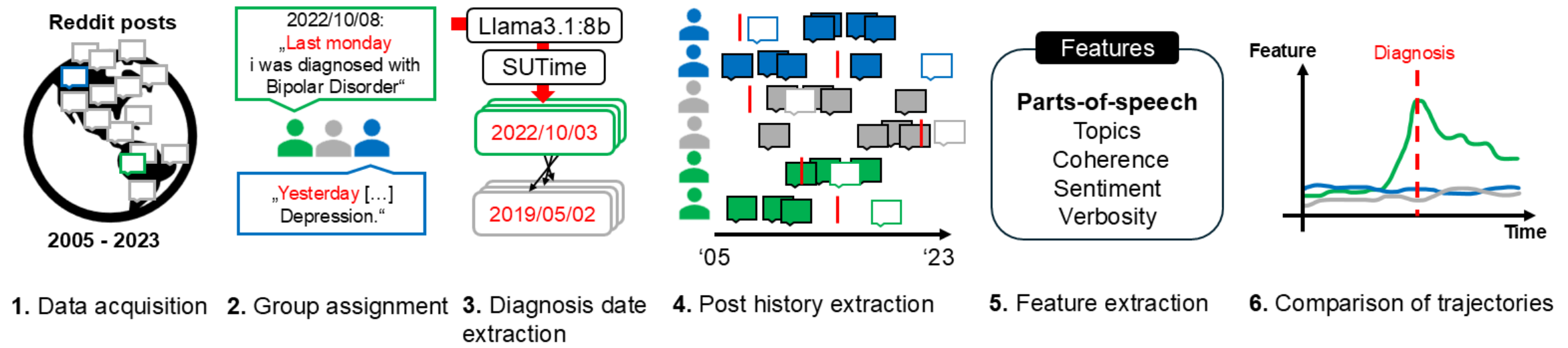

**Figure 1.** Overview of the study. We first extracted all Reddit comments and submissions between 2005 and 2023 from a freely accessible online repository. Users who self-disclosed a bipolar disorder (BD) or a depression (UD) are identified through regular expressions. Diagnosis self-disclosures were passed to a large language model (*llama3.1:8b*), which was prompted to extract time information on the diagnosis. This time information is then passed to *SUTime*, a temporal parsing algorithm, which yielded normalized datetime information. These data are additionally filtered through a rule-based algorithm to exclude non-viable datetimes (e.g., those including seasonal information such as "spring, 2022"). Pseudo-diagnoses were assigned to a group of regular Reddit users who served as a healthy control group (HC). The lifetime post history of all users was then extracted (2005-2023). Posts were analyzed using various natural language processing methods, such as neural topic modelling (Grootendorst, 2022), part-of-speech tagging, tokenization, semantic coherence, and sentiment analysis. Afterwards, trajectories of language features could be compared between groups.

## Methods

### Ethical considerations

All analyses included in this study were approved by the university's ethics committee (approval #1027). Although Reddit data is publicly available, mental health-related data is still sensitive. To protect users' privacy, we refrained from sharing usernames, uploading raw post data to public repositories, and verbatim-quoting posts (this ensures that users are not reverse-searchable on search engines).

### Reddit data

All available Reddit data between 2005 and 2023 was downloaded from a repository maintained by the Pushshift project (Baumgartner et al., 2020). Data from this source has been used extensively for research purposes (Wongkoblap et al., 2017). Download of compressed files took place in August 2024. Software/hardware used for this study is listed in Supplementary section 1.5.

### Clinical user identification

### Self-disclosure search

First, users who self-disclosed suffering from BD or UD were identified. Based on prior research, we defined six regular expressions that had to be contained in the submission or comment of a user for them to be identified as part of the BD or UD group (Birnbaum et al., 2017). Regular expressions are listed in the following: "diagnosed me with [DISORDER]", "diagnosed [DISORDER]", "i am diagnosed with [DISORDER]", "i was diagnosed with [DISORDER]", "i have been diagnosed with [DISORDER]", "my [DISORDER]", or "i have [DISORDER]", where [DISORDER] was exchanged with either "bipolar" (BD group) or "depression" (UD group). Overlapping users between BD and UD were excluded. Users whose account was deleted or whose username indicated that they were non-human agents (username contained "auto", "bot", "mod", or "admin") were excluded. In case of duplicate post texts, only the first was kept.

### Diagnosis information extraction

Self-disclosure posts were then passed to *Llama3.1-8b-Instruct*, an LLM, to retrieve any available time information on the diagnosis (Grattafiori et al., 2024). To this end, the following system prompt was designed: *"Your task is to extract the time of a [DISORDER] diagnosis in a text. Respond ONLY with the time of the diagnosis. If no other time information is present, respond with 'none'."*. The maximum number of response tokens was set to 10. The temperature parameter, which controls the randomness of responses, was set to 0 to maximize reproducibility (Demszky et al., 2023).

### Diagnosis time parsing

Diagnosis information was still in linguistic format (e.g., "yesterday") but a datetime format was needed (e.g., 2019/02/04). Passing LLM-retrieved time information and self-disclosure post timestamps to *SUTime* (Chang & Manning, 2012), a temporal parsing algorithm, returned diagnosis times in datetime format. A logical ruleset was devised which filters out unviable times returned by *SUTime* (see Supplementary section 1.1.).

### Control user identification

The strategy behind constructing the control user cohort was to find a collection of users which are representative of the general Reddit population but matches the BD group with respect to the distribution of post types (comments vs. submissions) and calendar time. To this end, a proportion-to-probability sampling strategy was followed, where the probability $P$ of retrieving a post in the month $m$ and of type $t$ from the database was formalized as

$$P(mt) = \frac{a_{mt} \times b}{n_{mt}} \quad (1)$$

where $a_{mt}$ is the proportion of all BD user posts falling into month m and post type $t$. $b$ is the number of total control posts to sample, which was set to 50,000. $n_{mt}$ is the number of posts in month $m$ and of post type $t$ that could be sampled (not empty, deleted, or removed). The resulting HC user identification posts were filtered according to the same criteria as the clinical users. HC users which overlapped with clinical users were removed. In cases of multiple identification posts for a given user, one post was randomly sampled. Pseudo diagnosis dates were randomly sampled with replacement from the distribution of diagnosis dates in the BD group.

**Post analysis**

All posts committed by the selection of users were extracted and analyzed. A list of criteria had to be fulfilled for a post to be eligible for analysis. Posts could not be empty strings, deleted, or removed. Posts could not contain any URL, which was checked by searching for a regular expression, and could not be enveloped by quotes or contain any text trailing a “>” string, which is a Reddit-specific symbol for quotes. Post containing URLs or quotes were highly likely to contain language produced by someone other than the posting user (Mangalik et al., 2024). Lastly, post had to be in the English language, which was determined using the “lingua” package.

Posts eligible for text analysis were first cleaned of Reddit-specific markup symbols using the “redditcleaner” package. Then, a variety of methods were applied to the posts to derive features reflecting the form and content of language.

**Formal language aspects**

Formal language features were derived from the literature on linguistic analysis of freely expressed speech in patients with psychosis or BD (Arslan et al., 2024; Corcoran et al., 2018; Parola et al., 2023). The number of words, sentences, and length of sentences was derived by tokenizing posts into words and sentences. The polarity of posts was derived through sentiment analysis based on VADER (Hutto & Gilbert, 2014). The polarity of a post, which ranges from -1 (very negative) to +1 (very positive), quantifies its emotional valence. The relative frequency of different syntax classes (e.g., proportion of personal pronouns), was determined by performing POS-tagging and normalizing the frequency of POS tags by the number of words. Analyses were implemented using the “textblob” package. Fig. 2c illustrates the process of computing coherence (Arslan et al., 2024; Parola et al., 2023; Plank & Zlomuzica, 2024b, 2024a).

**Language content**

Topic modelling was used to extract the content of posts (see Fig. 2a for an illustration). Neural topic modelling was chosen over Latent Dirichlet Allocation as the corpus contained many short posts (Egger & Yu, 2022; Grootendorst, 2022). Neural topic modelling followed the default pipeline of the “bertopic” package (Grootendorst, 2022). Only the top 250 most frequent topics were considered for further analysis (Lalk et al., 2024). To yield a topic score for each document and topic, the cosine similarity between the document and each topic embedding was calculated. A higher cosine similarity indicated that a given topic was more present in a document.

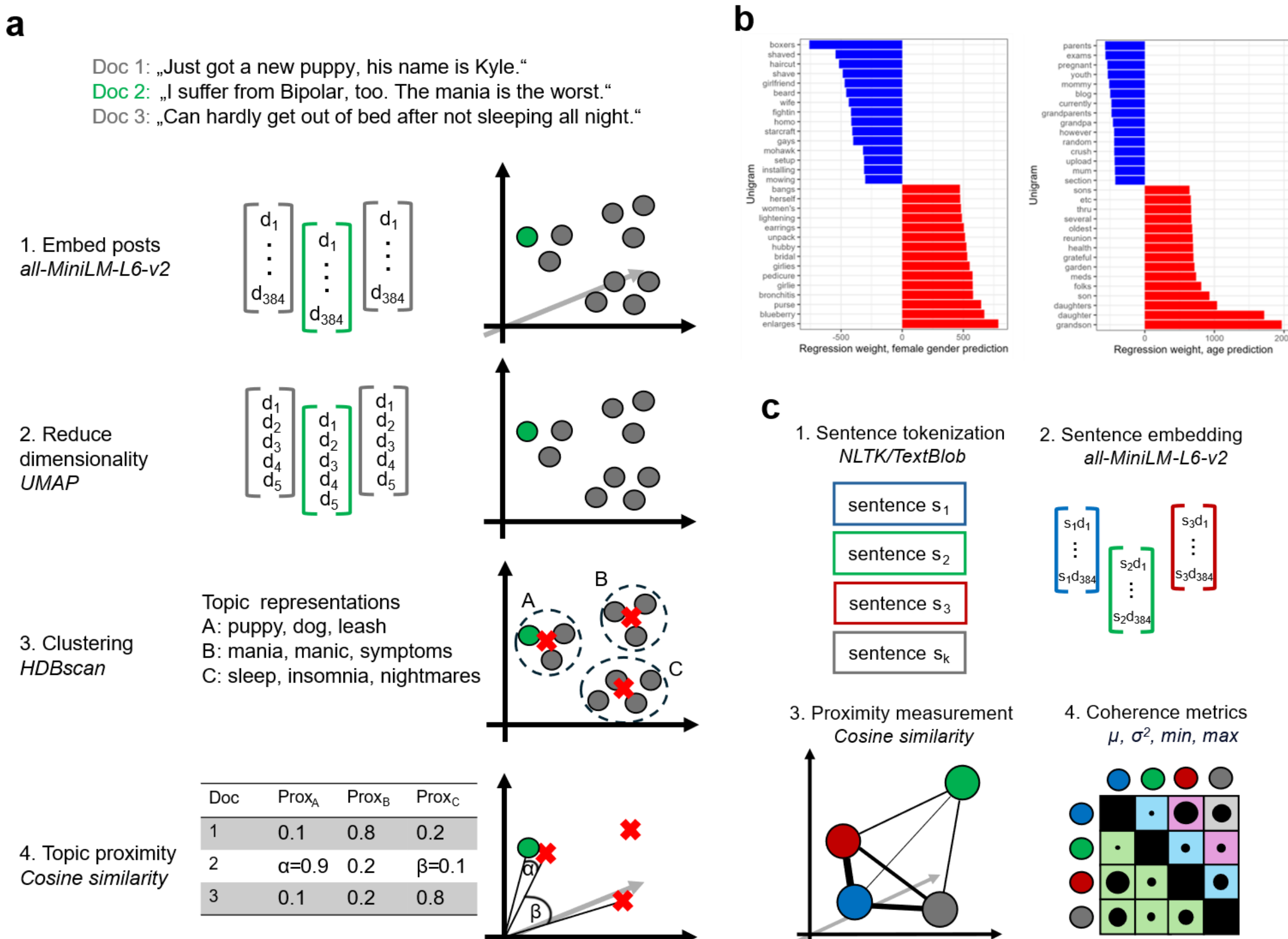

**Figure 2.** Natural language processing methods used in this study. **a**, Neural topic modelling with BERTopic (Grootendorst, 2022). Documents/posts are first transformed into semantic vectors using a sentence embedding model. The vectors' dimensionality is reduced through UMAP (McInnes et al., 2018). 5-dimensional semantic vectors are then clustered using HDBscan (McInnes et al., 2017). Through a class-based term-frequency inverse-document frequency procedure (Grootendorst, 2022), ten representational keywords can be extracted per cluster which allow for its interpretation as a topic. Topic embeddings are the average of constituent document embeddings. The cosine similarity of topic embeddings to a given document embedding measures the degree to which a topic is present within a given document. **b**, Regression weights for unigrams within the gender (left) and age (right) predictive lexica (Sap et al., 2014). The top 15 largest positive and negative regression weights are shown. In the gender lexicon, positive regression weights (marked red) indicate higher probability of female gender. In the age lexicon, positive regression weights (marked red) indicate older age. **c**, Coherence quantification (Parola et al., 2023). First, sentences were tokenized ("textblob" package) and embedded into semantic space (*all-MiniLM-L6-v2*). Then the cosine similarity of the embeddings of each sentence pair was calculated, yielding a similarity matrix. Three types of coherence can be calculated from the similarity matrix. The light-blue diagonal is the first-order coherence, calculated as the cosine similarity of each adjacent sentence vector pair. Second-order coherence, marked in magenta, is calculated as the cosine similarity of each sentence vector pair at an inter-sentence distance of 1 (every other sentence). Global coherence, marked as light green, is calculated as the cosine similarity of all possible sentence vector pairs. For all three coherence types, four statistical properties are calculated, namely the mean, variance, minimum, and maximum, yielding a total of 12 coherence features.

## Demography estimation

Age and gender information was derived by applying Sap et al.'s age- and gender-predictive lexica (Sap et al., 2014). We follow the same method in predicting our users' demographic data. Gender was first estimated based on each post individually. To yield singular gender values for each user, the majority class across all posts was determined to be the user's gender (female gender was ascertained in case of ties).

### Data aggregation and exclusion

Language features were aggregated for each month relative to the diagnosis date. To ensure high data quality among analyzed months, we excluded months with the lowest amount of data (i.e., those months at the outer edges of the distribution). Cutoffs were defined via $2.5^{th}$ and $97.5^{th}$ percentile as calculated in the group with the fewest users (UD). As a result, data from 45 months prior to the diagnosis to 253 months after the diagnosis was included.

### Data curation for periodicity analyses

In the post-diagnosis period, the three groups differed with respect to the average number of posts per user-month. Since this confounds periodicity analyses, we devised a stratified sampling procedure to ensure equal average number of posts per user between groups for each month (see Supplementary section 1.4). After this procedure, there were no differences in the average number of posts per user between groups (all *P*s > .1, Supplementary Fig. S1).

### Statistical analysis plan

Statistical analyses were performed in *R*.

### Diagnosis-sensitivity of language

The first objective was the discovery of language features which are sensitive to the onset of a BD diagnosis. To this end, time was categorized into a pre-diagnosis ($-45 \leq m \leq -7$) and acute period ($0 \leq m \leq 6$). Linear mixed models (LMMs) with each z-standardized language features as a criterion and a random user intercept were fit to the data ("nlme" package, *optim* optimization method, *nlminb* in case of convergence problems) using restricted maximum likelihood. The factors *Time* (pre-diagnosis vs. acute) and *Group* (0 = BD) served as predictors. Interactions of *Group* x *Time* informed us about differences in the change from pre-diagnosis period to acute period between groups. We first tested for *Time* x *Group* interaction effects on estimated age and gender. For gender as the criterion, a generalized LMM with a logit link function and binomial response variable distribution was fit ("lme4" package, bobyqa optimizer). *Age* and *Gender* were included as nuisance variables in subsequent models in case of significance of main or interaction effects ($\alpha = 0.05$). Number and length of sentences in posts were included as nuisance variables for all LMMs except for those with token features as criterion. Inspection of model fit revealed evidence of heteroscedasticity which was accounted for by including a residual weighting function based on the *Time* x *Group* interaction. The FDR was controlled for on a family-wise basis using the Benjamini-Hochberg's method (Benjamini & Hochberg, 1995) where families of tests were defined as those testing for group differences in topics, coherence, polarity, syntax, and verbosity, respectively.

### Language periodicity after diagnosis

The second objective was to analyze the periodicity of language features in the years following the diagnosis. This periodicity was analyzed both on a group-aggregate level using autocorrelation, and on an individual level using Lomb-Scargle Periodograms (LSPs). We considered data from 1 month to 253 months after the inferred diagnosis.

**Autocorrelation**. Pearson correlations of the *z*-standardized first derivative of averaged time series with itself at lag *L*, where $2 \leq L \leq 18$, were computed. Distributions of autocorrelations were derived for each of the three groups' averaged time series through a bootstrap procedure over 1,000 iterations with replacement (users, not monthly data were sampled) (Rousselet et al., 2023). 95% confidence intervals could be derived from the empirical quantiles of the resulting distribution (Rousselet et al., 2023). *p*-values of one-sided significance tests (BD > UD > HC) were defined as the proportion of bootstrap samples for whom the difference between groups' autocorrelation was ≥ 0 (Rousselet et al., 2023).

**Lomb-Scargle Periodograms.** Individual time series were sparse and unequally spaced which motivated the use of LSPs (Lomb, 1976; Scargle, 1982; VanderPlas, 2018). LSPs originate from astronomical research (Scargle, 1982) but have also been applied to the study of biological rhythms (Fonseca et al., 2013; Glynn et al., 2006; Ruf, 1999) and cosine regressions more generally have been used to study cyclical change in mood symptoms (Kiesner et al., 2016). LSPs are constructed by fitting candidate sine models to the observed data. The $\chi^2$-statistic of the best-fitting sine wave of a given frequency then serves as an estimate of the power of the signal at that frequency (VanderPlas, 2018). The maximum power of LSPs can be used as a measure of the periodicity of a time series. See Supplementary section 1.3. for more detail.

We compared the maximum power of LSPs between groups while controlling for confounding characteristics of the time series. Included confounders were the mean, median, variance, minimum, and maximum of the lengths of sequences of uninterrupted data, the number months that contained data, and the number of posts contributing to monthly data. The percentage of available data was also added as a confounder.

## Results

The pipeline returned a total of 60,454 Reddit users ($N_{BD}$ = 9,164, $N_{UD}$ = 6,173, $N_{HC}$ = 45,117). Times of diagnoses were inferred correctly in 75% of cases (Supplementary section 1.2.). Supplementary Fig. S1 and Table S2 provide descriptive statistics for the posts used to identify clinical users and infer their diagnosis. Users' posting histories comprised a total of 265,900,045 posts of which 174,592,914 (66,66%) were eligible for analysis.

We first aimed to gather insight into the topics that BD users discussed. To this end, we performed topic modelling which returned 244 interpretable topics. Fig. 2 shows a map of these topics. There were a variety of topics, some of which were mental health-related, while others were related to politics, social relations, animals, or recreational activities, and much more. The topic map captures semantic relations between topics: topics which are more proximal in semantic space are semantically more closely related. For example, the topic "eggs" (south of the map) connects food- and animal-related topics. A full listing of topics and their representative keywords is provided in Supplementary Table S1.

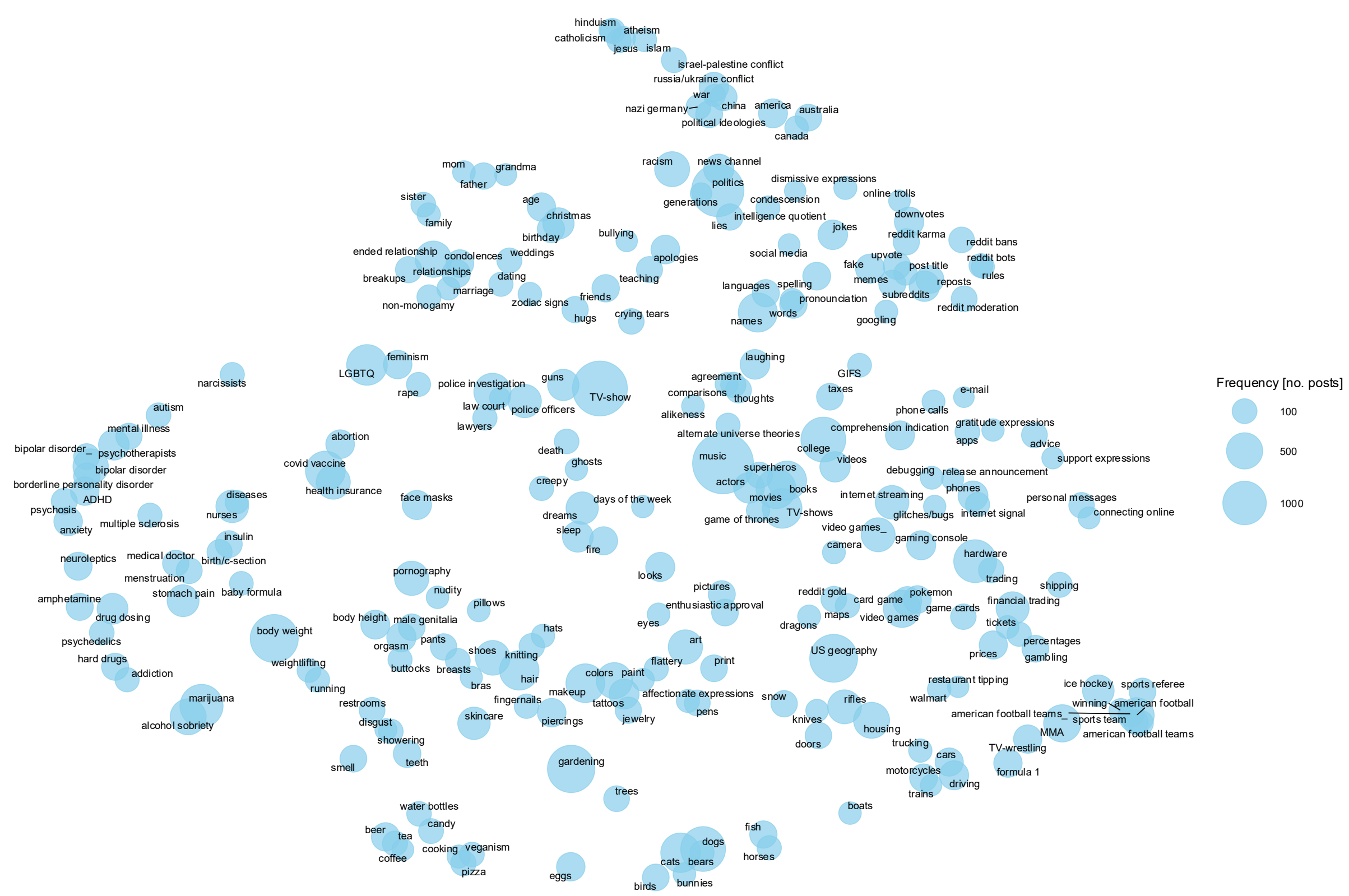

**Figure 3**. Topic map resulting from neural topic modelling on 202,044 posts from the group of bipolar disorder users (Grootendorst, 2022). Topic embeddings were reduced to two dimensions using UMAP. Spatial relations of labelled topics capture their semantic relationship. For example, the topic "eggs" (south on topic map) connects a cluster of animal-related topics (further east) with a cluster of food-related topics (further west). Light blue circles are scaled according to the number of documents that were assigned to a topic cluster.

## Language is sensitive to a bipolar disorder diagnosis

The goal of the first analysis was to determine whether there are changes to the language of BD users that coincide with the time of the self-disclosed diagnosis.

When checking for potential confounding demographic differences between groups, we found that BD users' have a higher log-likelihood of being female when compared to both HC ($B$ = -2.72, $SE$ = 0.3, $z$ = -8.97, $P$ < .001) and UD users ($B$ = -2.72, $SE$ = 0.3, $z$ = -2.46, $P$ = .014). The change in log-likelihood of female gender from pre- to post-diagnosis did not differ between groups (both $P$s ≥ .8).

BD users were also estimated to be older than both HC ($\beta$ = -0.06, $SE$ = 0.03, $t$ = -2.45, $p$ = .014) and UD users ($\beta$ = -0.14, $SE$ = 0.04, $t$ = -4.01, $P$ < .001). Additionally, the mean age increase from pre- to post-diagnosis was higher in the BD group when compared to both the HC ($\beta$ = -0.1, $SE$ = 0.03, $t$ = -3.59, $P$ < .001) and UD ($\beta$ = -0.11, $SE$ = 0.04, $t$ = -2.9, $P$ = .0038) group. Consequently, both estimated age and gender were added as nuisance variables in statistical models.

## Changes in language content

Significant *Time* x *Group* effects were found for 200 of 244 (81.97 %) topics in for the BD vs. HC comparison (visualized in Fig. 4d) and for 96 of 244 (39.34 %) topics for the BD vs. UD comparison. These indicate that the change from pre-diagnosis to the acute phase (*Time*) differed between groups (*Group*). Absolute *z*-standardized coefficients of significant effects ranged from $\beta$ = 0.05 to $\beta$ = 1.08 for the BD vs. HC comparison and from $\beta$ = 0.07 to $\beta$ = 0.54 for the BD vs. UD comparison. For brevity, we will summarize those effects which were consistent across both comparisons (92, 37.77 %). For a full list of results, readers are referred to Supplementary file S1.

### Mental disorders

The largest increase for both group comparisons could be observed for the topic *bipolar disorder* ($\beta_{BD}$ = 1.08, $\Delta\beta_{HC}$ = -1.08, $\Delta\beta_{UD}$ = -0.54). In addition, there were increases in talks about other mental health disorders such as *borderline personality disorder* ($\beta_{BD}$ = 1.04, $\Delta\beta_{HC}$ = -1.03, $\Delta\beta_{UD}$ = -0.48), *psychosis* ($\beta_{BD}$ = 0.99, $\Delta\beta_{HC}$ = -0.99, $\Delta\beta_{UD}$ = -0.44), *mental illness* ($\beta_{BD}$ = 0.94, $\Delta\beta_{HC}$ = -0.93, $\Delta\beta_{UD}$ = -0.38), *attention deficit hyperactivity disorder* ($\beta_{BD}$ = 0.93, $\Delta\beta_{HC}$ = -0.93, $\Delta\beta_{UD}$ = -0.36), *anxiety* ($\beta_{BD}$ = 0.83, $\Delta\beta_{HC}$ = -0.84, $\Delta\beta_{UD}$ = -0.26), *autism spectrum disorder* ($\beta_{BD}$ = 0.61, $\Delta\beta_{HC}$ = -0.62, $\Delta\beta_{UD}$ = -0.26), *addiction* ($\beta_{BD}$ = 0.74, $\Delta\beta_{HC}$ = -0.74, $\Delta\beta_{UD}$ = -0.22), and *alcohol dependence/sobriety* ($\beta_{BD}$ = 0.62, $\Delta\beta_{HC}$ = -0.63, $\Delta\beta_{UD}$ = -0.19).

### Psychoactive substances

BD users further showed an significant increase in talks about psychoactive substances such as *amphetamines* ($\beta_{BD}$ = 0.96, $\Delta\beta_{HC}$ = -0.96, $\Delta\beta_{UD}$ = -0.4), *neuroleptics* ($\beta_{BD}$ = 0.93, $\Delta\beta_{HC}$ = -0.93, $\Delta\beta_{UD}$ = -0.4), *psychedelics* ($\beta_{BD}$ = 0.81, $\Delta\beta_{HC}$ = -0.82, $\Delta\beta_{UD}$ = -0.35), *drug dosing* ($\beta_{BD}$ = 0.79, $\Delta\beta_{HC}$ = -0.79, $\Delta\beta_{UD}$ = -0.34), *hard drugs*, such as heroin or cocaine ($\beta_{BD}$ = 0.74, $\Delta\beta_{HC}$ = -0.72, $\Delta\beta_{UD}$ = -0.28), and *marijuana* ($\beta_{BD}$ = 0.51, $\Delta\beta_{HC}$ = -0.52, $\Delta\beta_{UD}$ = -0.19).

### Medical issues

BD users also talked more about medical issues such as *multiple sclerosis* ($\beta_{BD}$ = 0.86, $\Delta\beta_{HC}$ = -0.86, $\Delta\beta_{UD}$ = -0.3), *diseases* ($\beta_{BD}$ = 0.61, $\Delta\beta_{HC}$ = -0.59, $\Delta\beta_{UD}$ = -0.28), *COVID-19 vaccination* ($\beta_{BD}$ = 0.46, $\Delta\beta_{HC}$ = -0.43, $\Delta\beta_{UD}$ = -0.22), *insulin* ($\beta_{BD}$ = 0.68, $\Delta\beta_{HC}$ = -0.68, $\Delta\beta_{UD}$ = -0.21), and *birth/c-section* ($\beta_{BD}$ = 0.38, $\Delta\beta_{HC}$ = -0.39, $\Delta\beta_{UD}$ = -0.11)

### Somatic concerns

There were increases in topics related to somatic concerns such as *menstruation* ($\beta_{BD}$ = 0.58, $\Delta\beta_{HC}$ = -0.58, $\Delta\beta_{UD}$ = -0.2) and *stomach pain* ($\beta_{BD}$ = 0.63, $\Delta\beta_{HC}$ = -0.63, $\Delta\beta_{UD}$ = -0.19).

### Hospitalization

There were increases in topics related to hospitalization such as *nurses* ($\beta_{BD}$ = 0.68, $\Delta\beta_{HC}$ = -0.66, $\Delta\beta_{UD}$ = -0.27), *psychotherapists* ($\beta_{BD}$ = 0.81, $\Delta\beta_{HC}$ = -0.81, $\Delta\beta_{UD}$ = -0.26), *physicians* ($\beta_{BD}$ = 0.74, $\Delta\beta_{HC}$ = -0.75, $\Delta\beta_{UD}$ = -0.25), and *health insurance* ($\beta_{BD}$ = 0.37, $\Delta\beta_{HC}$ = -0.33, $\Delta\beta_{UD}$ = -0.09).

### Interpersonal issues

There were increases in topics concerning interpersonal issues such as *narcissists* ($\beta_{BD}$ = 0.43, $\Delta\beta_{HC}$ = -0.42, $\Delta\beta_{UD}$ = -0.19), *break-ups* ($\beta_{BD}$ = 0.24, $\Delta\beta_{HC}$ = -0.5, $\Delta\beta_{UD}$ = -0.14), *online trolls* (i.e., users who deliberately upset others) ($\beta_{BD}$ = 0.21, $\Delta\beta_{HC}$ = -0.18, $\Delta\beta_{UD}$ = -0.14), *family* ($\beta_{BD}$ = 0.5, $\Delta\beta_{HC}$ = -0.5, $\Delta\beta_{UD}$ = -0.14), *bullying* ($\beta_{BD}$ = 0.27, $\Delta\beta_{HC}$ = -0.26, $\Delta\beta_{UD}$ = -0.1), *relationships* ($\beta_{BD}$ = 0.31, $\Delta\beta_{HC}$ = -0.3, $\Delta\beta_{UD}$ = -0.1).

### Recreational activities

A group of topics was identified which were related to recreational activities. Some of these topics showed a stronger increase in the BD group relative to both controls. These were topics such as *TV-show*s ($\beta_{BD}$ = 0.08, $\Delta\beta_{HC}$ = -0.08, $\Delta\beta_{UD}$ = -0.14), *Pokémon* (cartoon franchise) ($\beta_{BD}$ = 0.03, $\Delta\beta_{HC}$ = -0.11, $\Delta\beta_{UD}$ = -0.06), *card games* ($\beta_{BD}$ = 0.03, $\Delta\beta_{HC}$ = -0.06, $\Delta\beta_{UD}$ = -0.11), *video games* ($\beta_{BD}$ = 0.05, $\Delta\beta_{HC}$ = -0.06, $\Delta\beta_{UD}$ = -0.09). Other topics related to recreational activities showed a decrease. These topics included *American football* teams ($\beta_{BD}$ = -0.09, $\Delta\beta_{HC}$ = 0.14, $\Delta\beta_{UD}$ = 0.09), *jewelry* ($\beta_{BD}$ = -0.31, $\Delta\beta_{HC}$ = 0.29, $\Delta\beta_{UD}$ = 0.11), *piercings* ($\beta_{BD}$ = -0.1, $\Delta\beta_{HC}$ = 0.08, $\Delta\beta_{UD}$ = 0.11), *tickets* ($\beta_{BD}$ = -0.35, $\Delta\beta_{HC}$ = 0.36, $\Delta\beta_{UD}$ = 0.15), *hats* ($\beta_{BD}$ = -0.36, $\Delta\beta_{HC}$ = 0.35, $\Delta\beta_{UD}$ = 0.15), *pants* ($\beta_{BD}$ = -0.31, $\Delta\beta_{HC}$ = 0.31, $\Delta\beta_{UD}$ = 0.17).

#### Ordinary experiences

There were decreases in many topics related to ordinary experiences. Some examples are the topics *cooking* ($\beta_{BD}$ = -0.24, $\Delta\beta_{HC}$ = 0.21, $\Delta\beta_{UD}$ = 0.12), *grocery shopping* ($\beta_{BD}$ = -0.25, $\Delta\beta_{HC}$ = 0.26, $\Delta\beta_{UD}$ = 0.11), and *cars* ($\beta_{BD}$ = -0.22, $\Delta\beta_{HC}$ = 0.23, $\Delta\beta_{UD}$ = 0.08).

#### Positive/hedonic experiences

We observed decreases in the prevalence of topics related to positive/hedonic experiences such as *winning* ($\beta_{BD}$ = -0.15, $\Delta\beta_{HC}$ = 0.17, $\Delta\beta_{UD}$ = 0.1), *memes* (i.e., internet jokes) ($\beta_{BD}$ = -0.18, $\Delta\beta_{HC}$ = 0.19, $\Delta\beta_{UD}$ = 0.07), *affectionate expression* (i.e., expression such as “This is so cute!”) ($\beta_{BD}$ = -0.17, $\Delta\beta_{HC}$ = 0.14, $\Delta\beta_{UD}$ = 0.1). or talks about Reddit-specific awards known as *gold* ($\beta_{BD}$ = -0.27, $\Delta\beta_{HC}$ = 0.24, $\Delta\beta_{UD}$ = 0.16).

#### Sexuality

Some of the pronounced decreases were found for topics related to sexually explicit topics such as *buttocks* ($\beta_{BD}$ = -0.37, $\Delta\beta_{HC}$ = 0.36, $\Delta\beta_{UD}$ = 0.22), *breasts* ($\beta_{BD}$ = -0.26, $\Delta\beta_{HC}$ = 0.25, $\Delta\beta_{UD}$ = 0.2), *nudity* ($\beta_{BD}$ = -0.29, $\Delta\beta_{HC}$ = 0.28, $\Delta\beta_{UD}$ = 0.15), and *male genitalia* ($\beta_{BD}$ = -0.15, $\Delta\beta_{HC}$ = 0.15, $\Delta\beta_{UD}$ = 0.1). In contrast, for the topic *orgasm* both comparisons were non-significant ($p_{FDR} \geq 0.26$), while for the topic *pornography* the BD group showed a significantly larger increase only when compared to the HC group ($\beta_{BD}$ = 0.12, $\Delta\beta_{HC}$ = -0.12).

#### Other notable topics

Other notable effects, which we could not be mapped to overarching themes, were increases in the topics *non-monogamy* (polygamy, polyamory, etc.) ($\beta_{BD}$ = 0.29, $\Delta\beta_{HC}$ = -0.27, $\Delta\beta_{UD}$ = -0.18), *Catholicism* ($\beta_{BD}$ = 0.21, $\Delta\beta_{HC}$ = -0.23, $\Delta\beta_{UD}$ = -0.17), *lawyers* ($\beta_{BD}$ = 0.27, $\Delta\beta_{HC}$ = -0.25, $\Delta\beta_{UD}$ = -0.16), *LGBTQ* ($\beta_{BD}$ = 0.31, $\Delta\beta_{HC}$ = -0.3, $\Delta\beta_{UD}$ = -0.16), and *dreams* ($\beta_{BD}$ = 0.42, $\Delta\beta_{HC}$ = -0.45, $\Delta\beta_{UD}$ = -0.12).

#### Unusual thought content

Lastly, there were three topics, that might reflect deviations from ordinary thought content, for which the BD group showed a greater increase than both the UD and HC group. These topics were *ghosts and paranormal phenomena* ($\beta_{BD}$ = 0.03, $\Delta\beta_{HC}$ = -0.06, $\Delta\beta_{UD}$ = -0.13), *zodiac signs/astrology* ($\beta_{BD}$ = 0.34, $\Delta\beta_{HC}$ = -0.37, $\Delta\beta_{UD}$ = -0.13), and a topic which we termed *alternate universe theories* ($\beta_{BD}$ = 0.15, $\Delta\beta_{HC}$ = -0.17, $\Delta\beta_{UD}$ = -0.12). The following paraphrased post provides an example of posts assigned to the latter topic.

*“To me, the omniverse represents an endless spectrum of realities and experiences that can be explored in countless ways. People can enter through dreams, stories, or video games, but I approach it through physics models—using them to envision worlds within this one, shaped by alternate rules of matter and different ways of interaction.”*

The user also appeared invested in his theory, stating (paraphrased):

*"I really wish I could share some pictures here—they’d at least help show the work I put into exploring the multiverse and give some validity to what I’m working on […].”*

In Supplementary section 2.2. we discuss to what extent these changes might be related to psychotic symptomatology.

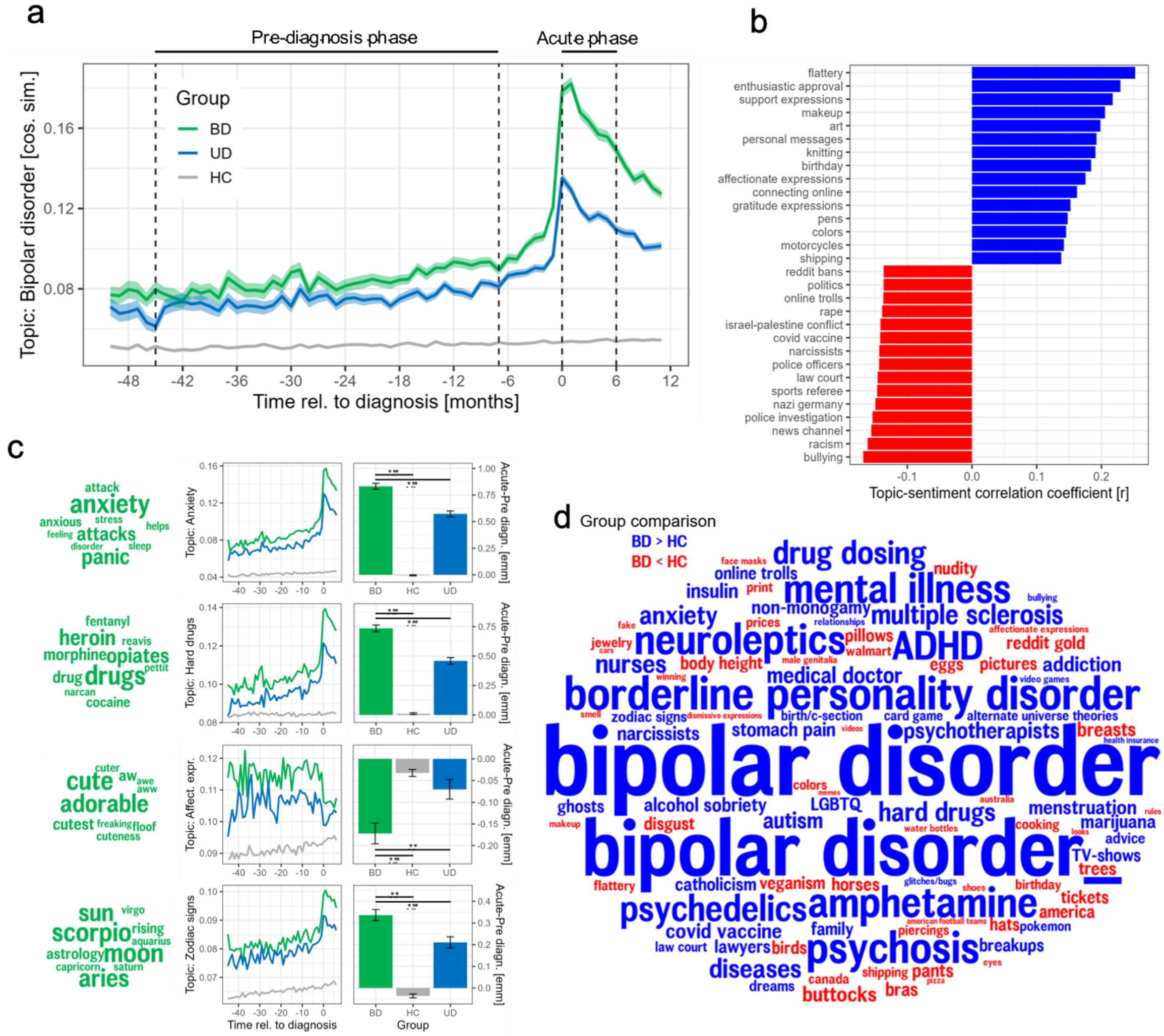

**Figure 4.** Diagnosis sensitivity of language. **a**, Monthly topic proximity for the topic bipolar disorder relative to the time of the diagnosis separately for each group. Time was dichotomized into a pre-diagnostic and acute period. Linear mixed models were used to compare the change from pre-diagnosis to acute phase between groups. **b**, The 30 topics that were most strongly correlated with posts' sentiment polarity. Correlations were computed from a random sample of 300,000 posts from three months in 2022. **c**, Topic representations (left column), raw trajectories (middle column), and change from pre-diagnosis to acute (right column) for four topics. From top-to-bottom, the topics were labelled *anxiety*, *hard drugs*, *affectionate expressions*, and *zodiac signs* (astrology). **d**, Word cloud displaying significant topic effects for the BD vs. HC group comparison. Topic sizes are scaled according to the absolute standardized coefficients of the *Time* x *Diagnosis* interaction. Blue topics are those where the increase from pre-diagnosis to acute was larger in the BD group, red topics indicate a stronger decrease from pre-diagnosis to acute in the BD group.

### *Sentiment analysis*

Both BD ($\beta$ = 0.09, *SE* = 0.03, *t* = 2.94, *P* = .003) and UD users ($\beta$ = 0.07, *SE* = 0.03, *t* = 2.14, *P* = .032) showed a stronger decrease in emotional valence than HC users. There was no significant difference in the effect of Time between BD and UD users ($\beta$ = 0.02, *SE* = 0.04, *t* = 0.57, *P* = .6. Therefore, posts by BD and UD users appeared to become equally more negative during the 6 months following the diagnosis.

Sentiment analysis could also be used to determine the emotional valence associated with different topics (Fig. 4d and Supplementary Table S1).

## Formal language change

### *Coherence*

Coherence is a common NLP-derived measure of language disorganization (Bilgrami, 2022; Parola et al., 2023) and can be defined as the flow of content in language (Bilgrami, 2022). Nine out of twelve coherence features showed a significant difference between the BD and HC group in the change from pre- to post-diagnosis. The absolute size of these effects ranged from $\beta = 0.08$ to $\beta = 0.21$. There was a pattern of increased average global coherence ($\beta = -0.1$), increased variance of coherence ($-0.18 \geq \beta \geq -0.21$), decreased minimum coherence ($0.081 \leq \beta \leq 0.083$), and increased maximum coherence ($-0.16 \geq \beta \geq -0.18$). Only the average first-order ($P_{FDR} = .2$) and second-order coherence ($P_{FDR} = .2$), and the minimum second-order coherence ($P_{FDR} = .08$) did not show a significant effect. We found no significant differences in the change of coherence features from pre- to post-diagnosis between the UD and BD group (all $P_{FDR} \geq .8$).

### *Verbosity*

BD users, when compared to HC users, showed an increase in the number of words ($\beta = -0.35$, $SE = 0.03$, $t = -11.31$, $P_{FDR} < .001$), number of sentences ($\beta = -0.32$, $SE = 0.03$, $t = -10.32$, $P_{FDR} < .001$), and sentence length ($\beta = -0.14$, $SE = 0.02$, $t = -8$, $P_{FDR} < .001$). There were no significant differences when BD users were compared to UD users (all $P_{FDR} \geq 0.1$).

### *Syntax*

When compared to HC users, BD users showed significant decreases in the relative frequency of determiners ($\beta = 0.14$, $P_{FDR} < .001$), existential *there* ($\beta = 0.09$, $P_{FDR} = .015$), singular or mass nouns ($\beta = 0.09$, $P_{FDR} = .008$), singular proper nouns ($\beta = 0.15$, $P_{FDR} < .001$), and significant increases in the relative frequency of personal pronouns ($\beta = -0.15$, $P_{FDR} < .001$), adverbs ($\beta = -0.16$, $P_{FDR} < .001$), the word *to* ($\beta = -0.09$, $P_{FDR} = .015$), and non-3rd person singular present verbs ($\beta = -0.17$, $P_{FDR} < .001$). There were no significant differences between BD and UD users (all $P_{FDR} = .1$).

## Language fluctuates periodically after the diagnosis

Upon inspection of raw linguistic trajectories, we noticed a visually discernable 12-month seasonality in the post-diagnosis period in both the BD and UD group (see Fig. 5a). This pattern is similar to yearly seasonal fluctuations of affect expression in Twitter posts found in an earlier study (Dzogang et al., 2016). We therefore sought to determine whether the observed 12-month seasonality would be statistically significant and increased in the BD group.

## Group-level periodicity

We first probed for differences in the 12-month autocorrelation of the *bipolar disorder* topic between groups. There was a significantly increased 12-month autocorrelation in both the BD ($r = 0.38$, *95% CI* = [0.28; 0.47], $P < .001$) and UD group ($r = 0.29$, *95% CI* = [0.15; 0.4], $P < .001$) relative to the HC group ($r = -0.02$, *95% CI* = [-0.19; 0.14]). There was no significant difference between the BD and UD group ($P = .13$). The increased autocorrelation in clinical groups was selective for the 12-month cycle and not observed for other cycle lengths between 2 and 18 months (Fig. 5b).

We then extended the autocorrelation analysis to all other topics. A total of 6 topic features showed a higher autocorrelation at lag 12 in the BD than in the HC group. The topics were borderline personality disorder ($P_{FDR}$ < .001), anxiety ($P_{FDR}$ < .001), bipolar disorder [first topic] ($P_{FDR}$ < .001), attention deficit hyperactivity disorder ($P_{FDR}$ < .001), bipolar disorder [second topic] ($P_{FDR}$ < .001), and psychosis ($P_{FDR}$ < .001).

The autocorrelation at lag 12 of the UD group was significantly greater than that of the HC group for 5 features after controlling the FDR. The topics were bipolar disorder [first topic] ($P_{FDR}$ = .049), attention deficit hyperactivity disorder ($P_{FDR}$ = .049), mental illness ($P_{FDR}$ < .001), psychosis ($P_{FDR}$ = .049), and bipolar disorder [second topic] ($P_{FDR}$ = .049).

There were no significant differences between the BD and UD group ($P_{FDR}$ > .9).

Evidently, both BD and UD groups' post topics exhibited an increased periodicity relative to HC users. Effects emerged exclusively for talks about psychopathology-related topics and were selective for the 12-month cycle (see Supplementary file S2 for full results).

**Seasonal profiles of BD diagnosis events and talks**

The discovered dominant 12-month cycle might be explained by well-established seasonal patterns in mood symptoms and diagnoses (Geoffroy et al., 2014). Although we lack information on users' geolocation, we reasoned that, since most Reddit users are from the US (Statista, 2024), seasonal profiles might still be recovered.

We first investigated seasonal profiles of self-reported diagnoses of affective disorders. Only self-disclosures which specified at least the month of the diagnosis were included in this analysis. Fig. 5c shows the seasonal profile of self-disclosed diagnoses of 6,960 BD and 4,650 UD users against monthly expected proportions (in a standard non-leap year). $\chi^2$ goodness-of-fit tests revealed that diagnoses were significantly different from monthly expected proportions in the UD ($\chi^2$ = 33.51, *df* = 11, *P* < .001), but not BD group ($\chi^2$ = 18.84, *df* = 11, *P* = .064). Post-hoc proportion tests revealed a significantly higher proportion of UD diagnoses falling into November (prop. = 0.094, expected = 0.082, $\chi^2$ = 8.7, $P_{FDR}$ = 0.038) and December (prop. = 0.094, expected = 0.085, $\chi^2$ = 10.8, $P_{FDR}$ = .024). Other comparisons were non-significant after FDR control (see Supplementary Table S5 for full results).

We then compared seasonal profiles of discussions of bipolar disorder symptoms. To this end we compared user-wise z-standardized *bipolar disorder* topic scores between groups for each month of the year. EMMs were derived from a LMM with a random user effect and fixed interacting and main effects of *Month* and *Group*. There were increased topic scores in the UD group compared to the HC group in January (*β = 0.049, SE = 0.015, t = 3.21,* $P_{FDR}$ *= .012*). In July, the UD group showed decreased topic scores relative to the HC group (*β = 0.062, SE = 0.015, t = 4.18,* $P_{FDR}$ *= .001*). The BD group, when compared to the HC group, showed increased topic scores in October (*β = 0.046, SE = 0.013, t = 3.54,* $P_{FDR}$ = .007) and November (*β = 0.043, SE = 0.013, t =3.35,* $P_{FDR}$ = .01). Other comparisons were non-significant after FDR control (see Supplementary Table S6 for full results).

In sum, these analyses provided evidence of seasonal profiles in both diagnoses of affective disorders and discussions of mood symptoms.

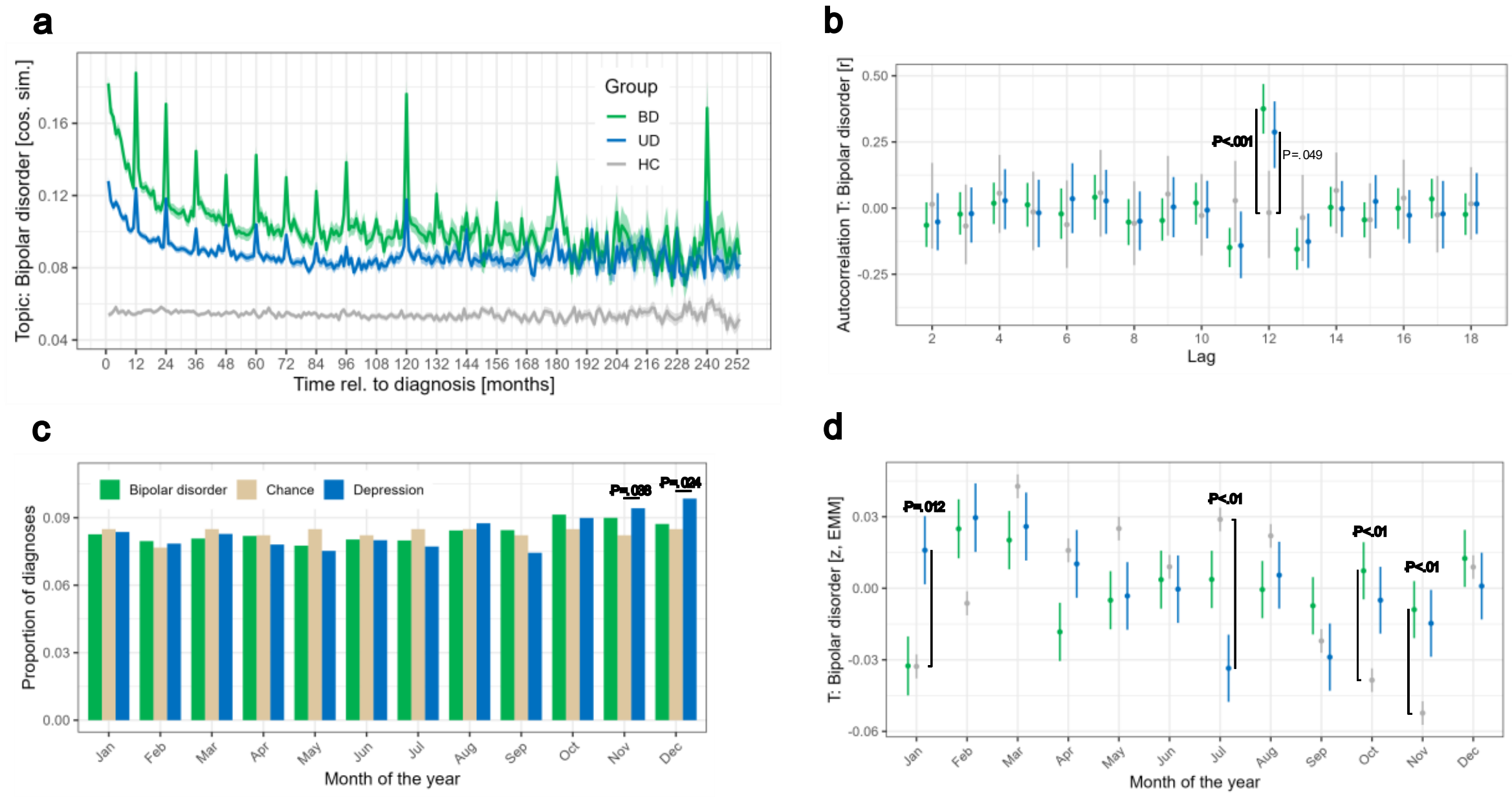

**Figure 5.** Language periodicity in the post-diagnosis phase. **a**, Raw monthly aggregates of the bipolar disorder topic in the post-diagnosis period for the three groups. **b**, Autocorrelation (Pearson's r) at different lags (months) for the bipolar disorder topic. Shown estimates are the means and 95% confidence intervals of bootstrap distributions. One-sided group comparisons revealed a significantly higher autocorrelation at lag 12 in the BD ($P_{FDR}$ < .001) and UD group ($P_{FDR}$ = .049) relative to the HC group. This effect was selective for the 12-month lag. **c**, Proportion of self-disclosed diagnoses falling into the different months of the calendar year. Only those self-disclosures were included where users identified at least the month of the event ($N_{BD}$ = 6,960, $N_{UD}$ = 4,650). Diagnoses were significantly different from monthly expected proportions in the UD (*P* < .001), but not BD group (*P* = .064). UD diagnoses were significantly more likely to fall into November ($P_{FDR}$ = 0.038) and December ($P_{FDR}$ = .024). **d**, Prevalence of the bipolar disorder topic for each group and month (shown as EMMs ± SE of user-wise z-standardized values). UD users showed an increase in topics scores in January ($P_{FDR}$ = *.012)* and a decrease in July ($P_{FDR}$ = .001*)* relative to HC users. BD showed increased topic scores in October ($P_{FDR}$ = .007*)* and November ($P_{FDR}$ = .01*)* compared to HC users.

### Individual-level periodicity

Next to seasonal variation of mood episode prevalence, patients' mood symptoms might adhere to their own individual frequencies. The goal of the following analysis was to quantify these individual-level periodicities. For this analysis, only users with at least 16 months of available data were considered (see Supplementary section 2.5.3.). Fig. 6 displays an LSP analysis for an exemplary UD user.

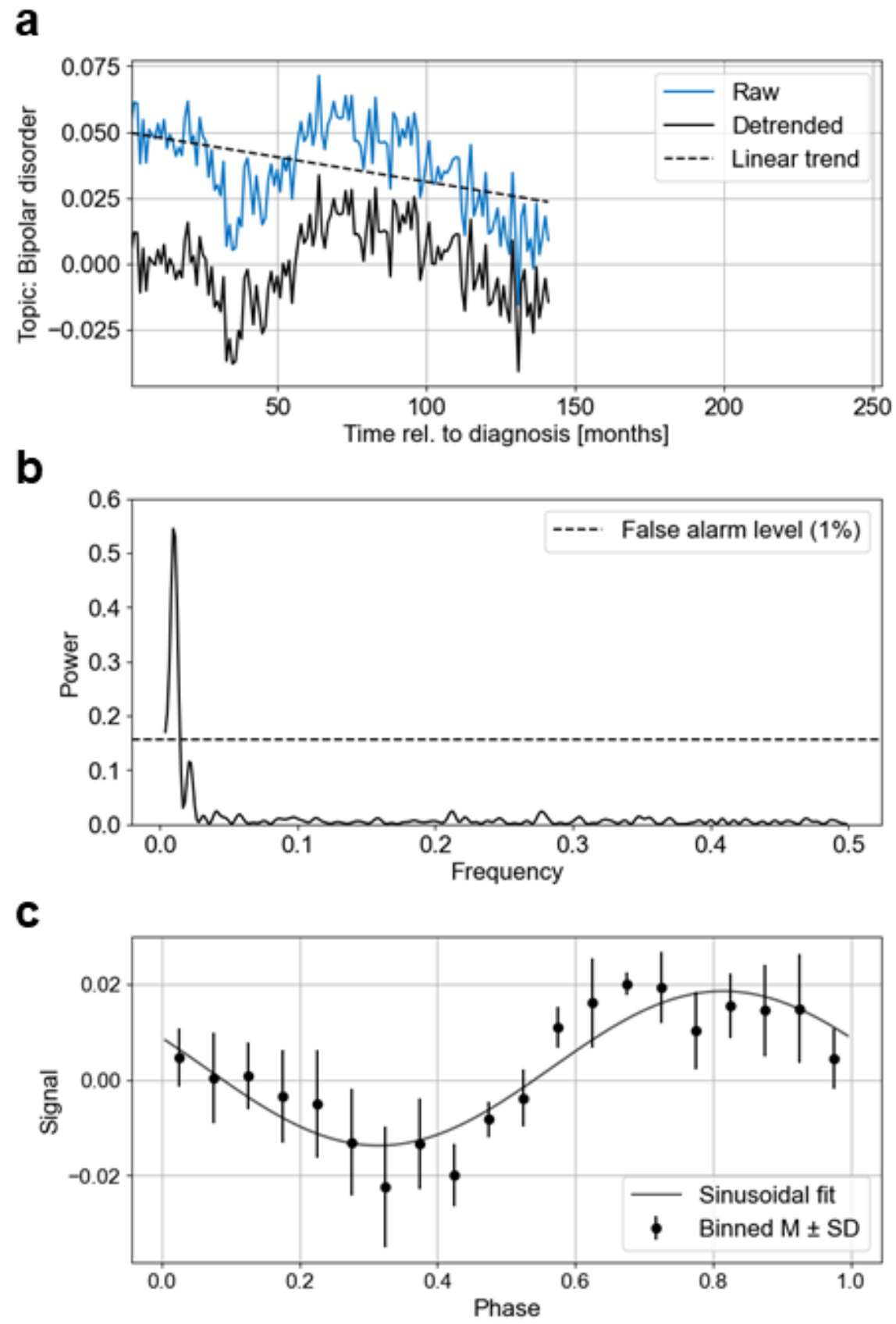

**Figure 6.** Individual-level periodicity analysis using Lomb-Scargle Periodograms (LSPs). **a**, shows the raw time series of the bipolar disorder topic for a single UD user (blue). Linear regression (dashed line) was used to detrend the signal (black line). **b**, shows the LSP for the detrended time series. There was a peak in power at a low frequency (0.008) which exceeded the false alarm level ($\alpha = 0.01$). **c**, the phase-folded signal with binned means ± SD and a sinusoidal fit for the dominant frequency. In Fig. S4, we provide an exemplary LSP of a time series which showed no significant peak.

Linear regression revealed significant group differences in the maximum LSP power. Pairwise group comparisons showed that both BD (*EMM* = 0.268, *SE* = 0.003, $P_{FDR} < .001$) and UD (*EMM* = 0.268, *SE* = 0.003, $P_{FDR} < .001$) users exhibited increased maximum power relative to HC users (*EMM* = 0.253, *SE* = 0.003). There was no difference between BD and UD users ($P_{FDR} = 1$). These differences were robust to the inclusion of estimated gender as a confounding variable which in itself had no significant effect on the maximum power ($B$ = 0.003, *SE* = 0.004, $t = 0.85$, $P = .4$).

While these results are in line with findings from the group-level, we note that peaks in LSPs were only significant for a minority of users (< 5%, see Supplementary section 2.5.3.). This means that while individual-level periodicity was significantly higher in clinical users when compared to unaffected users, for most time series we must assume that observed peak power is no different than what would be expected from a signal without a periodic component (VanderPlas, 2018).

## Discussion

Language alterations can be valuable markers of symptoms of BD, yet past investigations have been predominantly cross-sectional and of small scale (Arslan et al., 2024; Mota et al., 2014; Palaniyappan et al., 2019; Zaher et al., 2025). Here, we introduced a novel method to study language patterns longitudinally in SM users with BD. Our findings revealed pervasive changes to the content and form of language that mark the diagnosis of BD. In addition, we found patterns of periodic dsicussions about bipolar disorder symptoms in the years following a BD diagnosis. This periodicity was marked by a dominant 12-month cycle and could be related to mood fluctuations across seasons of the year.

Observed changes in language content map strikingly well to the clinical presentation of people with BD (Association & others, 2013). BD users increasingly talked about mood symptoms (American Psychiatric Association, 2022), psychiatric comorbidities (Krishnan, 2005), psychopharmacological agents (Bjørklund et al., 2016), illicit substances (Brown et al., 2001), hospitalization (Nierenberg et al., 2023), interpersonal conflict (Greenberg et al., 2014), sleep-related issues (Gold & Sylvia, 2016), medical comorbidities (Krishnan, 2005; Miller & Bauer, 2014), and issues related to the law (Lamberti et al., 2020). BD also users talked less about hedonic experiences and expressed less positive emotion, potentially hinting at depressive symptomatology such as anhedonia (Association & others, 2013; Strauss et al., 2016). These results contribute to discussions on the validity of using SM data for research into mental disorders (Birnbaum et al., 2017; Ernala et al., 2019; Plank & Zlomuzica, 2025). Indeed, the approach of determining clinical status from diagnostic self-disclosures has been criticized because self-disclosures could be inauthentic (i.e., irony, humor) (Birnbaum et al., 2017; Plank & Zlomuzica, 2025). The fact that linguistic changes map this well to the clinical presentation of BD indicates that the studied collection of users is actually suffering from BD thereby substantiating the idea that SM data can be used to generate valid insights into mental health disorders.

There was a pattern of periodic increases in discussions about mood symptoms in the years following the diagnosis. This periodicity was found for both BD and UD users and was consistent across group-level and individual-level analyses. The dominant 12-month cycle observed on a group level could be related to seasonal profiles of diagnoses and discussions about mood symptoms. In the UD group there was an increased proportion of diagnoses falling into November and December and increased talks of mood symptoms in January. In BD users, discussions of mood symptoms increased in September and October. These findings are in accord with an extensive literature on seasonal symptom and diagnosis patterns in affective disorders (Choi et al., 2011; Geoffroy et al., 2014; Øverland et al., 2020) and extend findings from an earlier study which demonstrated seasonal patterns of negative affect expression in non-clinical Twitter users (Dzogang et al., 2016). Because information on geolocation is not collected on Reddit, future studies could use the herein presented method to study mood periodicity from Twitter data (where geolocation can be shared (Mangalik et al., 2024)).

We also found changes in formal language aspects around the time of the diagnosis. These concerned the verbosity, syntax, and semantic coherence of posts in BD users. Findings on increased mean and variance of coherence are in line with a recent study of spoken language in patients with BD (Arslan et al., 2024) and provide further evidence that language disorganization might be evidenced from digital communications (Plank & Zlomuzica, 2024b). These effects were not specific to BD users which could be explained by heightened disorganization symptoms in patients with depressive disorders (Palaniyappan et al., 2025; Stein et al., 2025). Additionally, disorganization symptoms may be dependent on mood states (i.e., manic vs. depressed) (Palaniyappan et al., 2025; Zaher et al., 2025). We were unable to differentiate between mood states here which could have masked differences between BD and UD users. Differentiating between manic and depressive episodes will therefore be an important avenue for future research.

Syntactic changes, and in particular the observed increase in the use of personal pronouns, are in accord with evidence from previous research on spoken language in BD (Andreasen & Pfohl, 1976; Arslan et al., 2024; Lorenz & Cobb, 1952). In depression and psychosis, increases in personal pronoun usage have been taken as evidence of pathological self-focus and aberrant self-related processing (Edwards & Holtzman, 2017; Elleuch et al., 2025). Given the fact that both psychotic and depressive symptomatology can be present in BD (Grande et al., 2016; McIntyre et al., 2020), increases in personal pronoun use might therefore also exist in patients with BD.

Finally, there was an increase in the length of posts and sentences in BD users. The fact that this increase was also observed in UD users is at odds with findings of impoverished speech, such as poverty of speech, in depressed patients (Palaniyappan et al., 2025). This discrepancy probably indicates that verbosity in written language should not be equated to verbosity in spoken language. It is possible that increases in the length of posts – rather than indicating changes in verbosity – reflect an increased desire to seek social support from other SM users (Naslund et al., 2020).

SM data is highly naturalistic as it reflects unconstrained behavior in day-to-day life (Harvey et al., 2022; Wongkoblap et al., 2017). This fosters the ecological validity of language research in BD but necessarily coincides with a lower degree of experimental control. Since SM posts are unprompted, the content of language in SM datasets might be more variant than language generated in response to standardized speech elicitation protocols in laboratory studies (Harvey et al., 2022; Murray, 1943). We found that post topics correlate with formal language aspects (see Supplementary section 2.3. incl. Fig. S2). Therefore, differences in language form between BD and HC users, to some extent, might be explained by differences in language content. While we provide an analysis of language form while controlling for language content (Supplementary section 2.4.), we note that such analysis cannot meaningfully address this issue. Since talk about mood symptoms is correlated with the experience of mood symptoms (Eichstaedt et al., 2018; Mangalik et al., 2024; Schwartz et al., 2014), controlling for language content leads to a control for symptoms. This probably explains why most of the formal language changes became non-significant when language content was controlled for. Consequently, there is a trade-off between internal and ecological validity when studying language from laboratory studies vs. SM records that should be considered when interpreting results.

Past mental health research using SM data has not considered the time relative to a diagnosis thereby conflating behavioral change across different disorder stages (i.e., prodromal, acute, chronic). The herein-proposed method for extracting diagnosis times from SM records therefore presents a substantial methodological advancement. Notably, the method's utility is not confined to affective disorders or mental disorders in general. Instead, it can be generalized to any case where behavioral alterations relative to a discrete self-reported event are to be studied.

## Conclusion

Longitudinal studies of affective disorders have been limited by small scale, low granularity of assessment, and retrospective reporting biases (Reuben et al., 2016; Stromberg et al., 2025). The results of this study lend support for the notion that research using SM data is a useful adjunct to traditional psychiatric studies (Harvey et al., 2022; Wongkoblap et al., 2017). The method we introduced opens new avenues for longitudinal research into real-world aberrant behavior on a massive scale.

## Code and data availability

This study analyzed publicly available data which was provided by the Pushshift project. Code and processed data used in this study are available from the corresponding author upon reasonable request.

## Abbreviations

BD = bipolar disorder

UD = depression

HC = healthy control

LLM = large language model

NLP = natural language processing

SM = social media

POS = part-of-speech

UMAP = uniform manifold approximation and projection

c-TF-IDF = class-based term-frequency inverse-document-frequency

LMM = linear mixed model

FDR = false-discovery rate

LSP = Lomb-Scargle periodogram

EMM = estimated marginal mean

## Author contributions

LP: Conceptualization, methodology, software, validation, formal analysis, investigation, data curation, writing – original draft, writing – review & editing, visualization, project administration. AZ: Resources, writing – review & editing, supervision, funding acquisition.

## Conflicts of interest

The authors report no conflict of interest.

## Acknowledgements

This work was supported by grant No. ZL 59/4-1 and 59/5-1 to A.Z. from the German Research Foundation (Deutsche Forschungsgemeinschaft). The funding organization had no role in any aspects of this study.

The authors would like to thank the maintainers of the Pushshift project for enabling this research by providing access to data. The authors also thank Julia Christine Klenke for her help in manually inspecting the pipeline's accuracy as well as Kayleigh Piovesan, Jan Heistermann, and Giuliano Groer for fruitful discussions.